\documentclass[10pt,twocolumn,letterpaper]{article}

\usepackage{cvpr}
\usepackage{times}
\usepackage{epsfig}
\usepackage{graphicx}
\usepackage{amsmath}
\usepackage{amssymb}

\cvprfinalcopy 

\def\cvprPaperID{2411} 

\ifcvprfinal\fi
\begin{document}

\title{End-to-end learning of keypoint detector and descriptor \\ for pose invariant 3D matching}

\author{Georgios Georgakis$^{1}$, Srikrishna Karanam$^{2}$, Ziyan Wu$^{2}$, Jan Ernst$^{2}$, and  Jana Ko{\v{s}}eck{\'a}$^{1}$\\
$^{1}$Department of Computer Science, George Mason University, Fairfax VA\\
$^{2}$Siemens Corporate Technology, Princeton NJ\\
{\tt\small ggeorgak@gmu.edu,\{first.last\}@siemens.com,kosecka@cs.gmu.edu}
}

\maketitle


\begin{abstract}

Finding correspondences between images or 3D scans is at the heart of many computer vision and image retrieval applications and is often enabled by matching local keypoint descriptors. Various learning approaches have been applied in the past to different stages of the matching pipeline, considering detection, description, or metric learning objectives. These objectives were typically addressed separately and most previous work has focused on image data.  This paper proposes an end-to-end learning framework for keypoint detection and its representation (descriptor) for 3D depth maps or 3D scans, where the two can be jointly optimized towards task-specific objectives without a need for separate annotations. We employ a Siamese architecture augmented by a sampling layer and a novel score loss function which in turn affects the selection of region proposals. The positive and negative examples are obtained automatically by sampling corresponding region proposals based on their consistency with known 3D pose labels. Matching experiments with depth data on multiple benchmark datasets demonstrate the efficacy of the proposed approach, showing significant improvements over state-of-the-art methods.


\end{abstract}


\section{Introduction}
\thispagestyle{empty}
Keypoint representations have been a central component of matching, retrieval, pose estimation, and registration pipelines. With the advent of approaches based on deep neural networks, global representations became pervasive in solving these type of problems as they can be trained in a straightforward way in an end-to-end fashion.  Their shortcomings are caused by occlusions, partial views or scenes that contain large amount of clutter. In case of local feature representations, deep learning has been also applied to the different stages of the matching pipeline, considering detection, description, or metric learning objectives. Most of the frameworks considered the above objectives separately, used image data, and required a large number of training examples. 
In order to mitigate these issues, we propose to use deep convolutional networks for learning keypoint representations and a keypoint detector for 3D matching jointly without the need for separate annotations. The costly annotation stage can be avoided due to the availability of large repositories of 3D models and the capability of obtaining depth images from different viewpoints.


\begin{figure}
\begin{center}
 \includegraphics[width=1\linewidth]{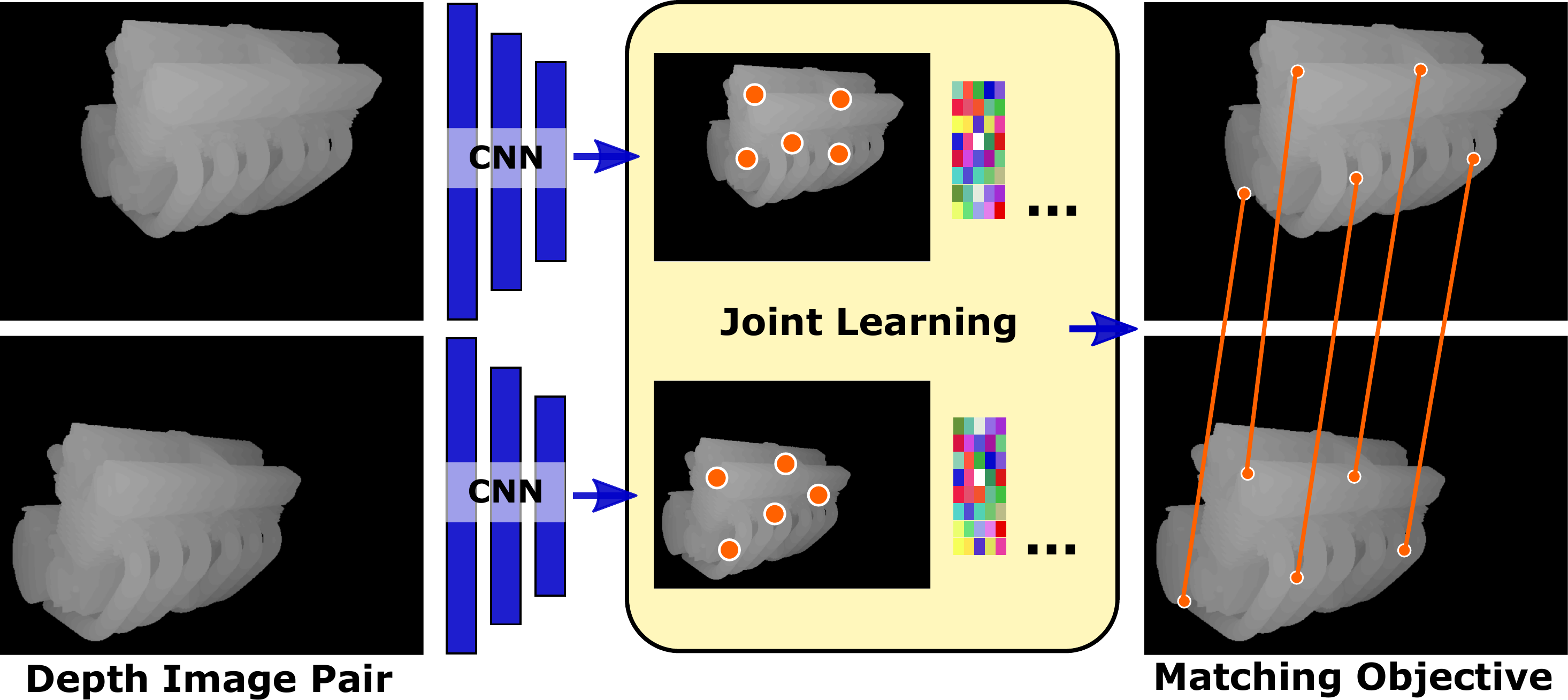} 
\end{center}
   \caption{We propose a new method for jointly learning keypoint detection and patch-based representations in depth images towards the keypoint matching objective.}
   \vspace{-0.3cm}
\label{fig:title_im}
\end{figure}


For the problem of jointly learning keypoint detectors and descriptors, we define a Siamese network architecture that receives as input a pair of depth images and their pose annotations. Each branch of the architecture is a proposal generation network used to generate patches in the two depth images. The branches share weights and lead to a sampling layer which selects pairs of patches. The pairs are labeled as positive or negative depending on the proximity of their 3D re-projection calculated from the pose labels. In other words, the sampling layer is used to create ground truth data on-the-fly by taking advantage of the initial pose annotations. For training the network, we use the contrastive loss which attempts to minimize the distance in the feature space between positive pairs, and maximize the distance between negative pairs. Therefore, for patches that are very close in the 3D space, but sampled from different images, we are learning a representation that has minimal distance in the feature space. 
In order to learn where to select patches from, we define a {\em score loss}  to gauge the performance of the target task. For example, for pose estimation the {\em score loss} should consider the number of positive matches between two images from different viewpoints. 
To summarize, the key contributions of this paper include the following:
\begin{itemize}
\setlength\itemsep{0em}
\item We propose the first end-to-end framework for joint learning of keypoint detector and local feature representations for 3D matching,
\item We propose a novel sampling layer that can generate labels for local patch correspondence on-the-fly, and 
\item We design a {\em score loss}  encapsulating task specific objectives that can implicitly provide supervision for joint learning of keypoint detector and its feature representation. 
\end{itemize}
We evaluate the matching accuracy of the proposed approach on multiple benchmark datasets and demonstrate improvements over state-of-the-art methods.


\begin{figure*}
\begin{center}
 \includegraphics[width=\linewidth]{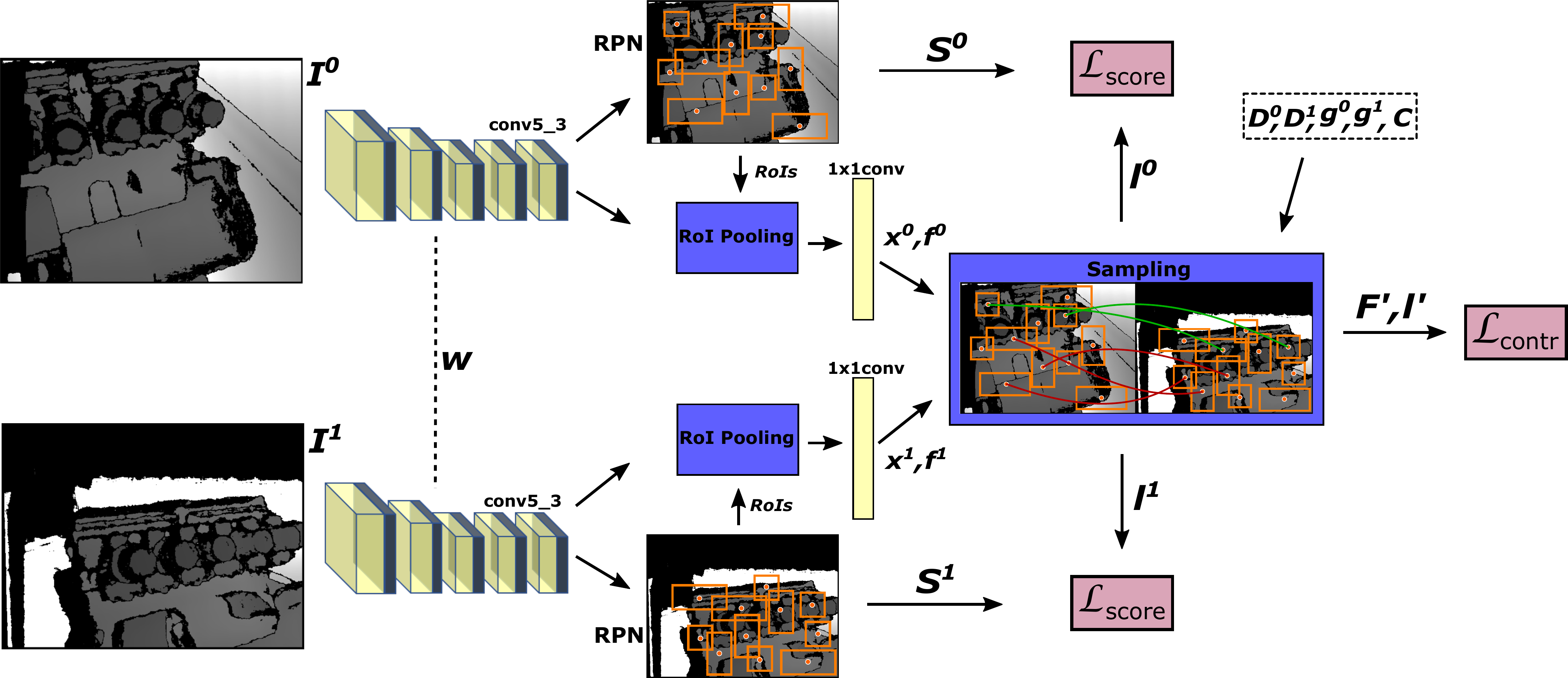} 
\end{center}
   \caption{Overview of our Siamese architecture. Each branch is a modified Faster R-CNN which receives as an input a depth image and uses VGG-16 as the base representation network. Features from conv5\_3 are fed into both the Region Proposal network (RPN) and the Region of Interest (RoI) pooling layer. Given a set of proposals from RPN, we pass their scores to the score loss, while their RoIs are fed to the RoI pooling layer and a fully connected layer to extract the feature vectors. The RoI centroids and the features from both branches are then passed to the sampling layer which organizes them into pairs used by the contrastive loss. Note that the weights between the two branches are shared. For more details on the notations please see section~\ref{sec:training}.}
\vspace*{-1.5em}
\label{fig:architect}
\end{figure*}

\section{Related Work}
There is a large body of work on keypoint detectors and descriptors for both images and 3D depth maps. For images, features such as SIFT~\cite{lowe2004distinctive}, FAST~\cite{rosten2010faster}, BRISK~\cite{leutenegger2011brisk}, and ORB~\cite{rublee2011orb} have been used effectively for various matching tasks. Detectors and descriptors specifically designed for 3D data, including feature histograms~\cite{rusu2008aligning} and geometry histograms~\cite{frome2004recognizing} are already included in the Point Cloud Library (PCL) along with many others~\cite{rusu_ICRA2011}. These representations were hand-engineered with specific keypoint matching accuracy and/or efficiency goals. A comprehensive review of 3D descriptors can be found in~\cite{guo_IJCV2016}.


Advances in convolutional networks led to works in learning descriptors and distance metrics for various matching tasks.  The descriptor learning problem has been extensively tackled in case of images and was typically formulated as a supervised learning problem. Given positive and negative examples of pairs of descriptors, the goal is to learn representations where the positive examples are nearby and negative examples are far apart. The methods vary between those which use fixed descriptors and learn a discriminative metric to approaches which take raw patches and learn new representations, or both. 3D reconstructions are often used to obtain large amounts of training data. A comprehensive evaluation of existing approaches can be found in~\cite{schonberger_CVPR2017}.

Most relevant to our task are the descriptor learning methods of Zagoruyko et al.~\cite{zagoruyko2015learning}, Han et al.~\cite{han2015matchnet}, and Wohlhart et al.~\cite{wohlhart2015learning}, where patch representations are learned discriminatively by means of Siamese or Triplet networks, considering pairs or triplets of descriptors. 
Similar approaches have been proposed for learning feature representations for matching 3D data~\cite{zeng20173dmatch,song2016deep,maturana20153d}. 
Both in case of images and depth maps, the feature descriptors were typically computed at fixed sized patches or patches determined by sampling both spatial locations and scale.

The problem of learning the detector was addressed by Salti et al.~\cite{salti_ICCV2015}, where a descriptor specific keypoint detector was proposed by casting the problem of selecting keypoint locations and spatial support as a binary classification task. Savinov et al.~\cite{savinov_CVPR2017} formulates the keypoint detection problem as the problem of learning how to rank points consistently over various image transformations.
Other methods such as~\cite{Yi_ECCV16, altwaijry2016learning, lenc_ECCV2016} rely on hand-crafted interest point detectors to collect training data, which is done separately from the training process, affecting the learning of the keypoint detector. In contrast to these approaches, we formulate the problem of selecting keypoints (locations and spatial support) and their feature representations in a single, unified framework, enabling joint optimization of the parameters for both.


\section{Approach}
We are interested in jointly learning a keypoint detector and a view-invariant descriptor using depth data. In contrast to other approaches (\cite{Yi_ECCV16}, \cite{salti_ICCV2015}), our work does not use hand-crafted keypoint detectors or descriptors as initialization for the learning procedure. Since it is unclear in case of 3D data which keypoint locations should be labeled as ``interesting'', we do not rely on any hand-labeled datasets with keypoint annotations. Instead, we use a modified Faster R-CNN~\cite{NIPS2015_5638} as the head of our architecture to bootstrap the learning process. 
Specifically, given two depth images with some pose perturbation, we first generate two sets of proposals, one for each image. Then, we project the proposals in 3D using the known image poses in order to establish positive and negative pairs. Proposals with a small distance in 3D are considered correspondences and are therefore labeled as positives. The pairs are then passed to a contrastive loss in an attempt to minimize feature distance between positive pairs and maximize the distance between negative pairs. Additionally, we introduce a new score loss, which finetunes the parameters of the Region Proposal Network (RPN) of the Faster R-CNN~\cite{NIPS2015_5638} to generate high-scoring proposals in regions of the depth maps for which we can consistently find correspondences. 
To the best of our knowledge, this is the first work that attempts to jointly optimize the keypoint detection and representation learning process in a purely self-supervised fashion.

\subsection{Architecture}
We choose to use Faster R-CNN~\cite{NIPS2015_5638} as the basis for our architecture because of its modularity. Even though it is initially trained for the task of object detection, its components can provide us with patch-based representations and a trainable mechanism for selecting those patches. 
We use Faster R-CNN as part of a Siamese model with shared weights. Both branches are connected to a layer responsible for finding correspondences which we call the sampling layer. A contrastive loss is used to train the representation and each branch has a score loss for training the keypoint detection stage. An overview of the architecture with more details can be seen in Figure~\ref{fig:architect}.




\subsection{Training}
\label{sec:training}
In order to train our model, we require pairs of depth images \{$I^0$, $I^1$\} each with its camera pose information \{$g^0$, $g^1$\} and the intrinsic camera parameters $C$. These can be obtained by rendering a 3D model from multiple viewpoints or using RGB-D video sequences with registered frames (\cite{schonberger_CVPR2016}, \cite{izadi2011}). To pass the depth images \{$I^0$, $I^1$\} through our network, we first normalize their depth values in the RGB range and replicate the single channel into a 3-channel image. The rest of the inputs $g^0$, $g^1$, $C$, and the depth images with their values in meters $D^0$, $D^1$ are passed directly to the sampling layer.

For each depth image, the Region Proposal Network (RPN) generates a set of scores, and regions of interest (RoIs) for which we use their centroids as the keypoint locations. Each RoI also determines the spatial extent used for feature computation for the current keypoint and after RoI pooling layer, we obtain the representation for each keypoint. We keep the top $t$ keypoints based on their scores and establish our set of keypoints, $K^m = \left\lbrace (\textbf{x}^{m}_{0}, s^{m}_{0}, f^{m}_{0}), ..., (\textbf{x}^{m}_{t}, s^{m}_{t}, f^{m}_{t}) \right\rbrace$, where $m=\left\lbrace0,1\right\rbrace$ corresponds to the pair of depth images, $\textbf{x}^{m}_{t} = (x_t,y_t)$ are 2D coordinates on the image plane, $s^{m}_{t}$ is the score which signifies the saliency level of the keypoint, and $f^{m}_{t}$ is the corresponding feature vector.

\begin{figure*}
\begin{center}
 \includegraphics[width=\linewidth]{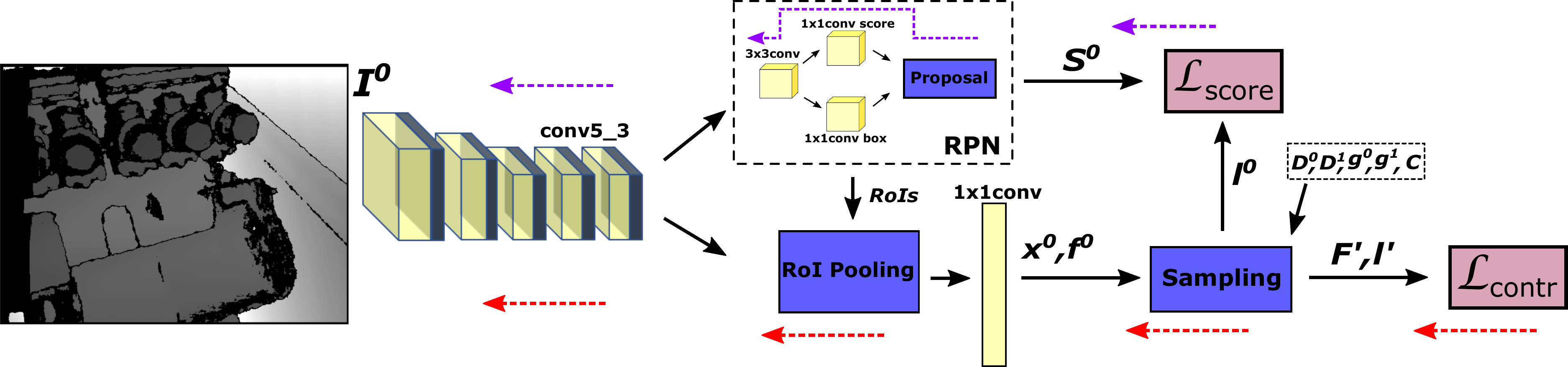}
\end{center}
   \caption{Gradient backpropagation during training of our network. The figure only shows one branch. The purple and red arrows show the path of the gradients from the score and the contrastive loss respectively. Notice that no gradients are passed in the 1x1 conv bbox layer, since we are not optimizing towards bounding box regression.}
   \vspace*{-1.5em}
\label{fig:gradient}
\end{figure*}

The sampling layer then receives the sets of keypoint centroids and their features from both images, \{$\textbf{x}^0, \textbf{x}^1, f^0, f^1$\}. To determine the correspondences between the keypoints of the two images, the centroids are first projected in 3D space. For each keypoint $\textbf{x}^{0}_i$, we find the closest $\textbf{x}^{1}_j$ in 3D space based on Euclidean distance and form the $n^{th}$ pair of features $F'_n = (f^0_i, f^1_j)$. If the distance is less than a small threshold, we label it as positive ($l'_n=l^0_i=l^1_j=1$), otherwise it is considered as a negative pair ($l'_n=l^0_i=l^1_j=0$). This can possibly lead to one class vastly outnumbering the other. However, this can be advantageous for learning the keypoints, as the number of positive pairs indicates how many keypoints were generated consistently between the two input depth images. This is different from the correspondence layer used in~\cite{schmidt_ICRA2017}, which performed dense sampling of correspondences, and had no notion of keypoints or their repeatability.

\textbf{Joint Optimization:} As mentioned earlier, we are interested in jointly learning a view-invariant representation along with a keypoint detector. Towards this end we introduce the following multi-task loss:
\begin{equation}
\begin{aligned}
L(\left\lbrace K^0 \right\rbrace , \left\lbrace K^1 \right\rbrace) &= \lambda_c L_c (F',l') + \\
& \lambda_s L_s^0(s^0,l^0) + \lambda_s L_s^1(s^1,l^1)
\end{aligned}
\end{equation} 
where, $L_c$ is a slightly modified contrastive loss which operates on the pairs of the keypoints and optimizes over the representation, $L^m_s$, are the score loss components which use the keypoint scores in order to optimize the detector,
$l'$ is the set of labels of the set of feature pairs $F'$, and $\lambda_c$ and $\lambda_s$ are the weight parameters. Note that since we formed the features into the set of pairs $F'$, we use the notation $n$ to signify the $n^{th}$ feature pair $(f^0_n, f^1_n)$. The contrastive loss is defined as:
\begin{equation}
\begin{aligned}
L_c (F',l') &= \frac{\sum_{n=1}^N l'_n||f_n^0-f_n^1||^2}{2N_{pos}} + \\
& \frac{\sum_{n=1}^N(1-l'_n)max(0, v-||f_n^0-f_n^1||)^2}{2N_{neg}} 
\end{aligned}
\end{equation}
where $v$ is the margin, and $N_{pos}$, $N_{neg}$ are the number of positive and negative pairs respectively ($N=N_{pos}+N_{neg}$).
Each class contribution to the loss was normalized based on its population to account for the imbalance between the positive and negative pairs. The score loss is defined as:
\begin{equation}
L_s^m(s^m,l^m) = \frac{1}{1+N_{pos}} - \frac{\gamma \sum_{i=1}^N l^m_i \log s_i^m}{1+N_{pos}}
\end{equation}
where $l^m_i$ is the label for the $i^{th}$ keypoint from image $I^m$ whose value depends whether the keypoint belongs to a positive or negative pair, and $\gamma$ is a regularization parameter. Note that since the pairs are formed by picking a keypoint from each image and each keypoint can belong to only one pair, then $|l^0|=|l^1|=N$.

The objective of the score loss is to maximize the number of correspondences  between two views.
We specifically avoid looking for discriminative keypoints as that would entail defining the meaning of a discriminative keypoint. This is ambiguous by nature as discriminativeness can be subjective, depended also on the task at hand. Instead, we consider ``interesting'' keypoints as those for which we can find correspondences between two viewpoints, and ideally we want RPN to rank them higher than others. Therefore, we optimize towards generating as many positive keypoints as we can, in addition to maximizing their scores. We consider only the positive pairs and penalize them if their generated score is low. The loss is normalized by the number of positives, however, $\gamma$ can be utilized to regulate the trade-off between optimizing for the number of keypoints versus optimizing for the scores.

Furthermore, our framework allows regulating the trade-off between number of matches and localization accuracy during training, by adjusting the 3D distance threshold in the sampling layer. For example, with a small threshold, the model will learn to associate few keypoints with high accuracy as opposed to a large number with a more relaxed threshold. Since we are generating annotations on-the-fly, this enables us to train systems with varying trade-off between matching likelihood and accuracy to address application needs.

During backpropagation, we pass the gradient for each keypoint at the appropriate location in the gradient maps, by storing their locations during the forward pass and implementing the backwards functionality in the region proposal layer. For the score loss, the gradients are passed through the convolutional layers that are responsible for predicting the scores. In contrast to the traditional Faster R-CNN, we do not finetune the bounding box regressor as there are no ground-truth boxes available for our task. However, our training scheme implicitly affects the bounding box generation, as all preceding layers are trained.
An illustration of how the gradients are backpropagated for both losses in one branch of our network can be seen in Figure~\ref{fig:gradient}. 

\begin{figure}[t]
\begin{center}
		\includegraphics[width=0.4\linewidth]{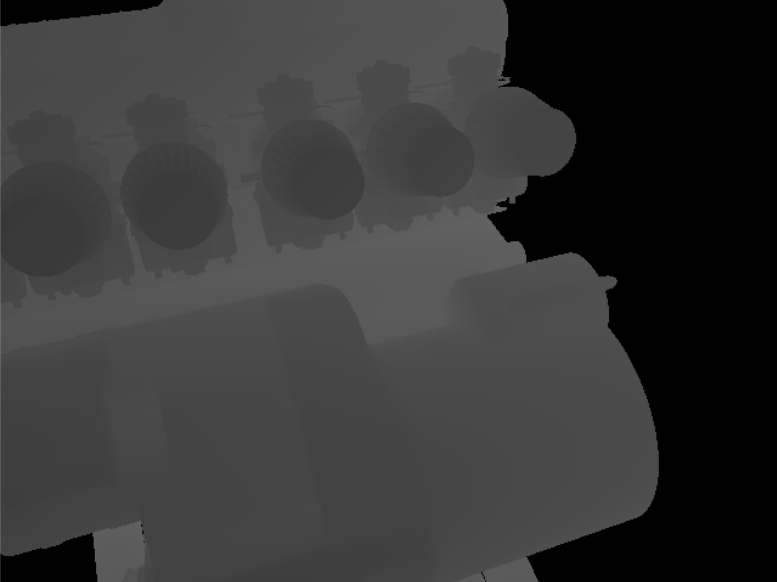}
 		\includegraphics[width=0.4\linewidth]{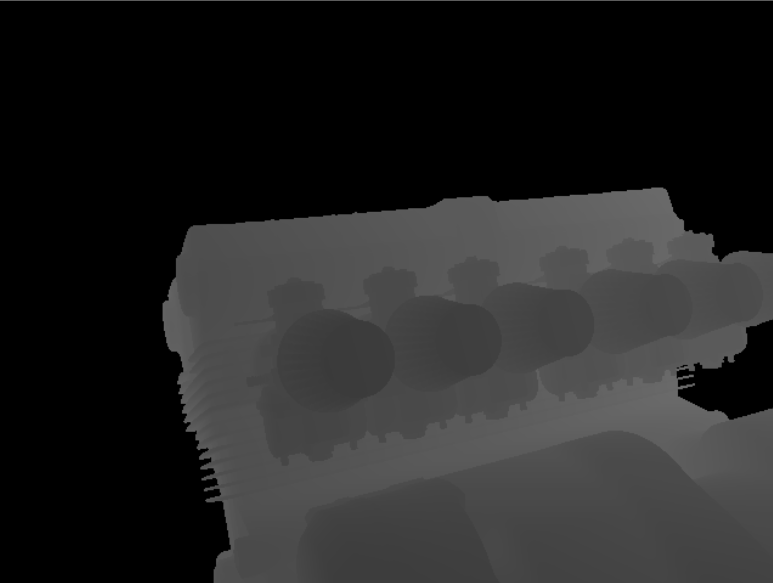}
 		\\
 		\vspace*{1mm}
		\includegraphics[width=0.4\linewidth]{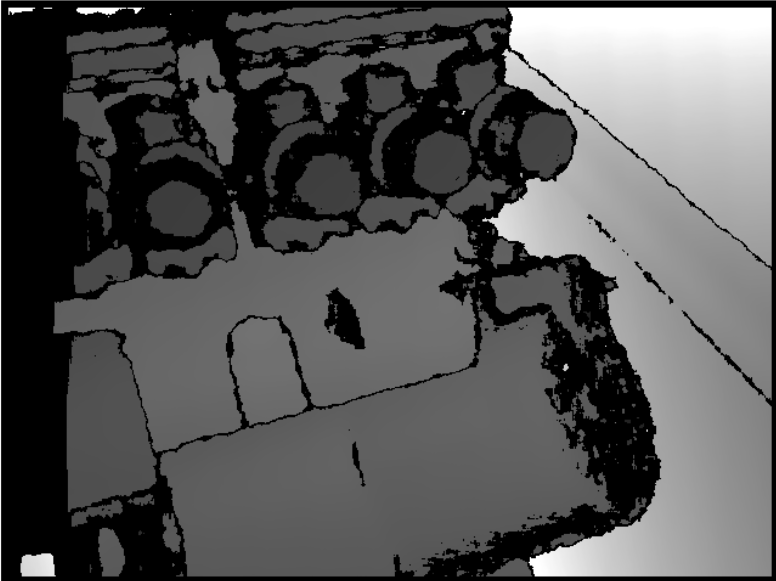}
		\includegraphics[width=0.4\linewidth]{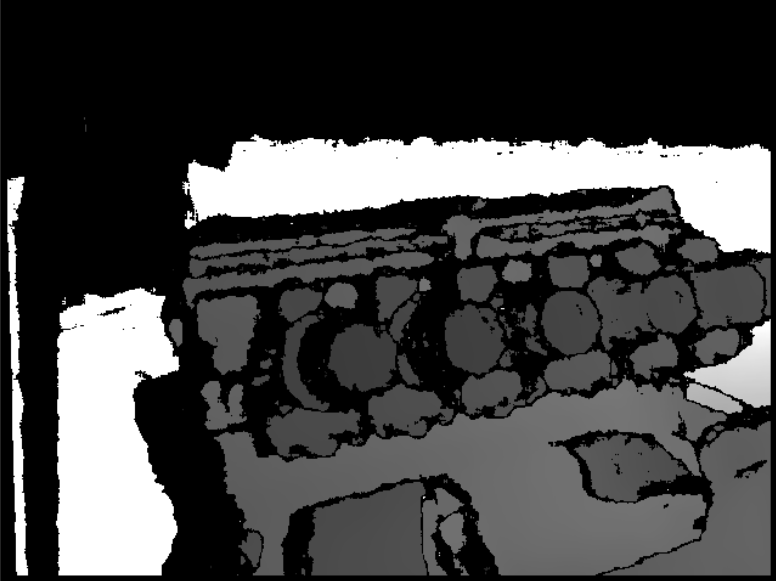}
\end{center}
   \caption{A training pair for the \textit{Engine} model shown here both noise-free (top row) and noisy (bottom row) created using DepthSynth~\cite{planche_arXiv2017}.}
   \vspace{-0.5cm}
\label{fig:model}
\end{figure}

\section{Experiments}
In order to validate our approach, we compare its matching capabilities to hand-crafted features, the keypoint learning method KPL~\cite{salti_ICCV2015}, and the state-of-the-art 3DMatch~\cite{zeng20173dmatch} which learns 3D local geometric descriptors using a siamese deep learning architecture. For the hand-crafted features we form 4 baselines from the combinations of the 3D keypoint detectors Harris3D~\cite{sipiran2011} and ISS~\cite{zhong_ICCV2009} and the 3D descriptors FPFH~\cite{rusu_ICRA2009} and SHOT~\cite{salti2014} found in the Point Cloud Library (PCL)~\cite{rusu_ICRA2011}. KPL~\cite{salti_ICCV2015} is a descriptor-specific keypoint learning approach for which we use the provided trained model. We combine it with the SHOT descriptor as it was proposed by the authors of~\cite{salti_ICCV2015}. Similarly, since 3DMatch is a local 3D descriptor, we combine it with Harris3D keypoint detector and use the model trained for keypoint matching provided by the authors. In addition, we add one more baseline which is a variation of our method, where we train using only the contrastive loss. We refer to this baseline as \textit{Ours-No-Score}. 


Two main experiments are performed. First, we test on a set of 3D models, both with clean and noisy data, and second, we evaluate on two datasets captured by a real depth sensor.
Our motivation for choosing these datasets is to compare the performance of our work with the baselines when dealing with noise from a depth sensor. Other works~\cite{salti_3DIMPVT2011} usually apply Gaussian noise on the 3D models to simulate the noise, however, this does not sufficiently represent realistic scenarios. Therefore, for the first experiment, we use DepthSynth~\cite{planche_arXiv2017}, which synthetically generates realistic depth data from 3D CAD models by modeling vital factors such as sensor noise, material reflectance, and surface geometry that affect the scanning process. 
The synthetic noisy images produced by DepthSynth are thus much closer to the real depth images output from structured light depth sensors.


\begin{figure}
\begin{center}
 \includegraphics[width=1\linewidth]{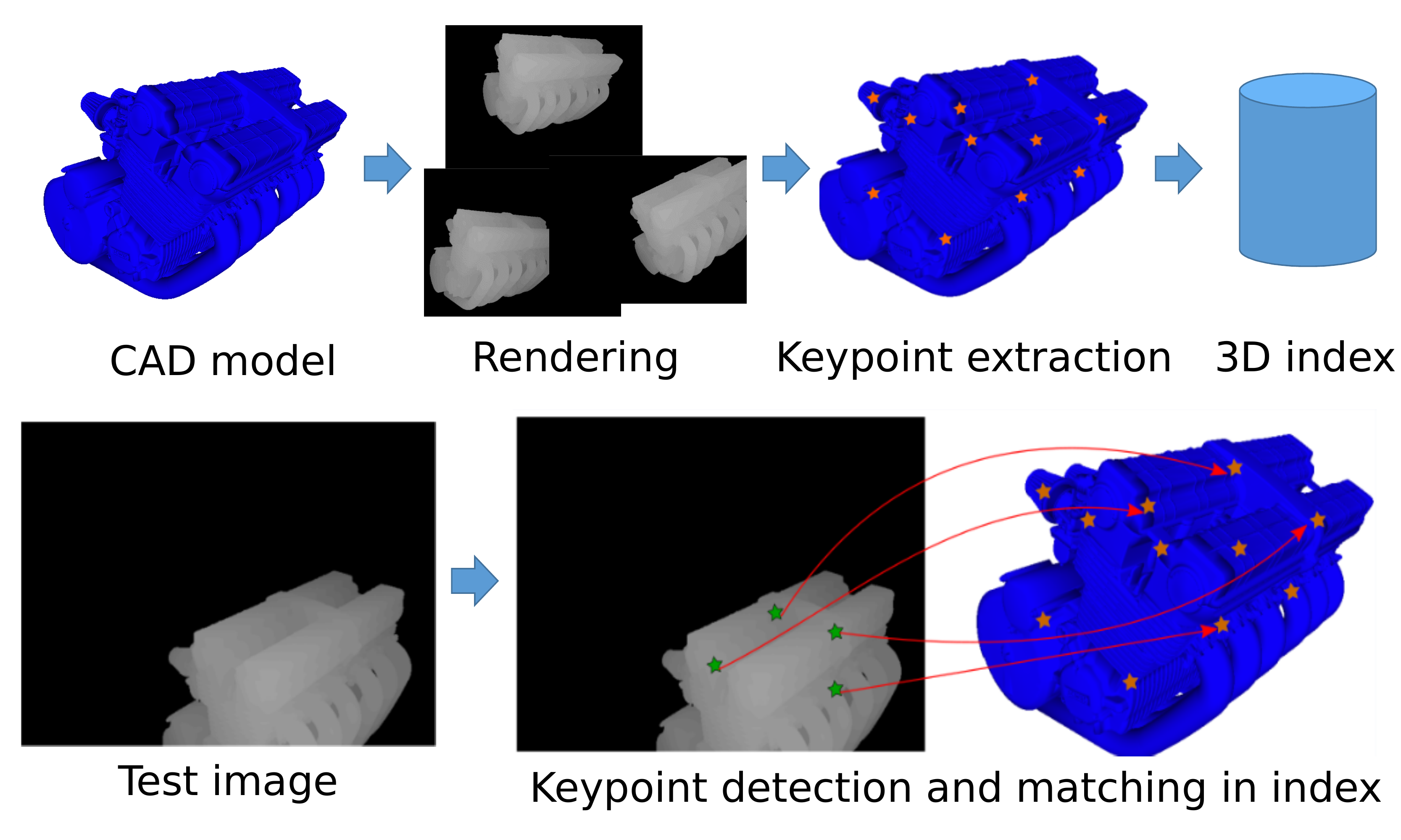}
\end{center}
   \caption{Overview of the evaluation pipeline used in our experiments. The top row describes the repository creation, while the bottom shows the test procedure.}
\label{fig:pipeline}
\end{figure}

\begin{table}[t]
\begin{center}
	\begin{tabular}{|c|c|c|}
	\hline
	Method & Noise-Free & Noisy \\
	\hline
	ISS~\cite{zhong_ICCV2009}+SHOT~\cite{salti2014} & 47.9 & 0.5 \\
	\hline
	KPL~\cite{salti_ICCV2015}+SHOT~\cite{salti2014} & 57.2 & 2.8 \\
	\hline
	ISS~\cite{zhong_ICCV2009}+FPFH~\cite{rusu_ICRA2009} & 61.1 & 2.9 \\
	\hline
	Harris3D~\cite{sipiran2011}+SHOT~\cite{salti2014} & 60.1 & 5.9 \\	
	\hline
	Harris3D~\cite{sipiran2011}+FPFH~\cite{rusu_ICRA2009} & \textbf{79.1} & 12.8 \\
	\hline
	Harris3D~\cite{sipiran2011}+3DMatch~\cite{zeng20173dmatch} & 66.2 & 20.7 \\		
	\hline
	\hline
	Ours-Rnd & 29.8 & 7.3 \\
	\hline
	Ours-No-Score & 40.7 & 11.1 \\	
	\hline
	Ours-Transfer & - & 17.8 \\
	\hline
	Ours	 & 67.4 & \textbf{23.8} \\
	\hline
  	\end{tabular}
    \caption{Keypoint matching accuracies (\%) comparison on both noise-free and noisy views from the \textit{Engine} 3D model.}
    \label{tab:res_engine}
\end{center}
\end{table}

\begin{figure}[t]
\begin{center}
 		\includegraphics[width=0.48\linewidth]{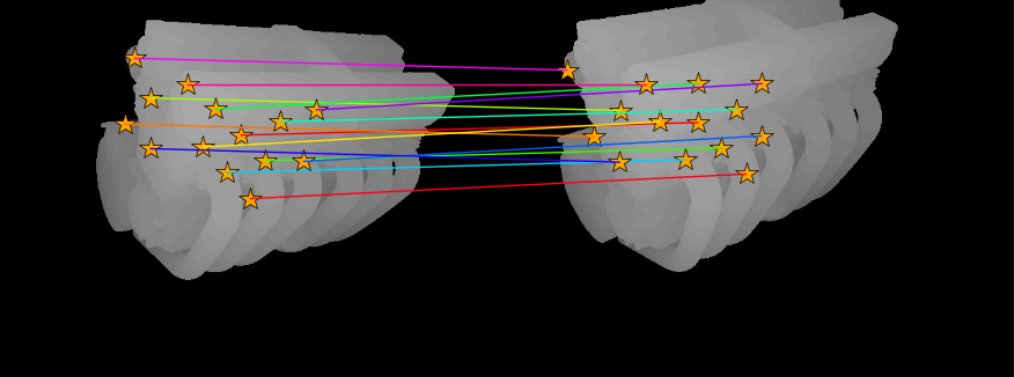}
		\includegraphics[width=0.48\linewidth]{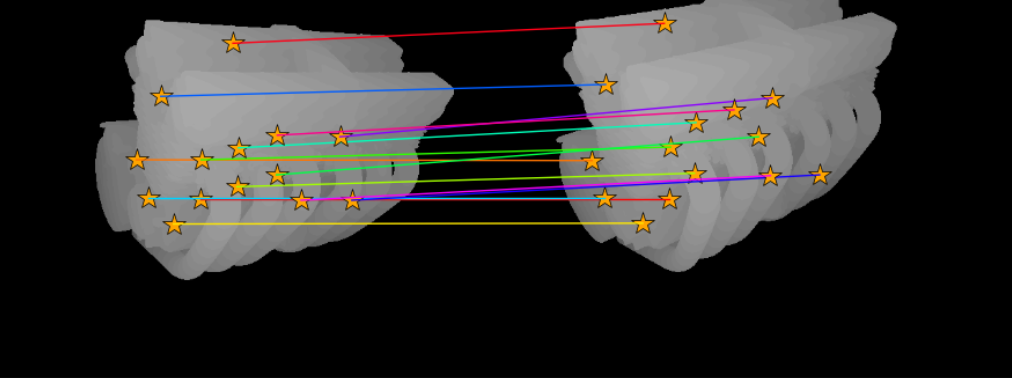}
		\\
		\vspace*{1mm}
		\includegraphics[width=0.48\linewidth]{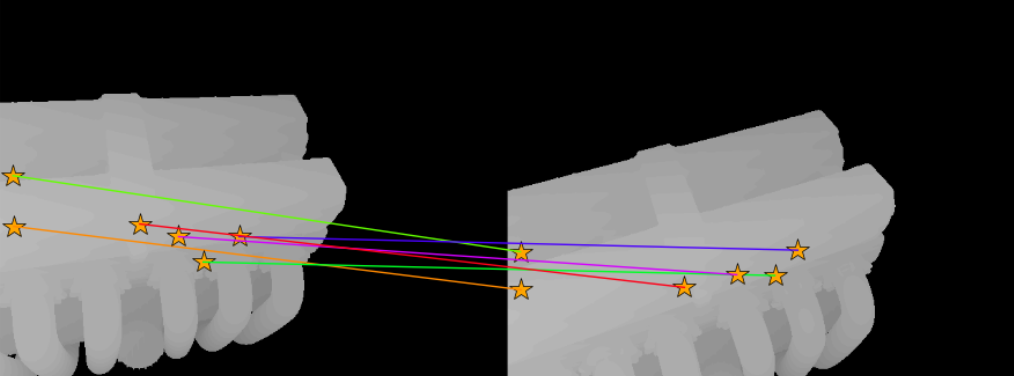}
		\includegraphics[width=0.48\linewidth]{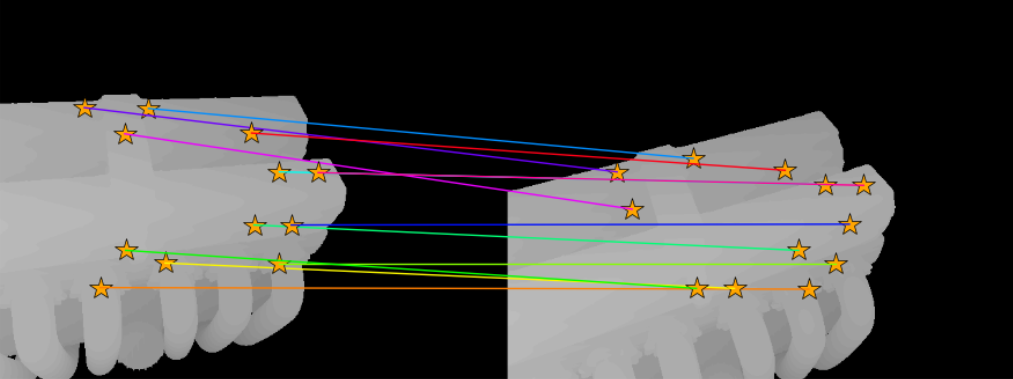}		
\end{center}
   \caption{Qualitative demonstration of the contribution of the score loss on matching examples on the noise-free views from the \textit{Engine} model. The examples where the model was trained without the score loss (left column) contain smaller number and less accurate matches in comparison to the examples with the model trained with the score loss (right column). Best viewed in color.}
\label{fig:no_score_matching}
\end{figure}



\begin{table*}[t]
\begin{center}
	\begin{tabular}{|c|c|c|c|c||c|}
	\hline
	Method & Armadillo & Bunny & Dragon & Buddha & Average \\
	\hline
	ISS~\cite{zhong_ICCV2009}+SHOT~\cite{salti2014} & 0.8 & 0.5 & 0.6 & 0.4 & 0.6\\
	\hline
	ISS~\cite{zhong_ICCV2009}+FPFH~\cite{rusu_ICRA2009} & 2.0 & 1.7 & 2.4 & 1.4 & 1.9\\
	\hline
	Harris3D~\cite{sipiran2011}+SHOT~\cite{salti2014} & 8.0 & 11.4 & 6.9 & 6.7 & 8.3\\
	\hline
	KPL~\cite{salti_ICCV2015}+SHOT~\cite{salti2014} & 18.0 & 12.8 & 15.4 & 9.1 & 13.8\\	
	\hline
	Harris3D~\cite{sipiran2011}+FPFH~\cite{rusu_ICRA2009} & 14.5 & 16.0 & 16.4 & 10.5 & 14.4\\
	\hline
	Harris3D~\cite{sipiran2011}+3DMatch~\cite{zeng20173dmatch} & 14.9 & 17.7 & 27.8 & 15.1 & 18.8 \\
	\hline
	\hline
	Ours-No-Score & 10.0 & 18.3 & 25.2 & 12.5 & 16.5\\
	\hline
	Ours	 & \textbf{25.2} & \textbf{31.9} & \textbf{45.7} & \textbf{27.7} & \textbf{32.6}\\
	\hline
  	\end{tabular}
    \caption{Keypoint matching accuracies (\%) comparison on noisy data from the Stanford 3D models.}  
    \label{tab:res_synthetic_noisy}
\end{center}
\end{table*}

\begin{figure*}[t]
\begin{center}
	\includegraphics[width=0.83\linewidth]{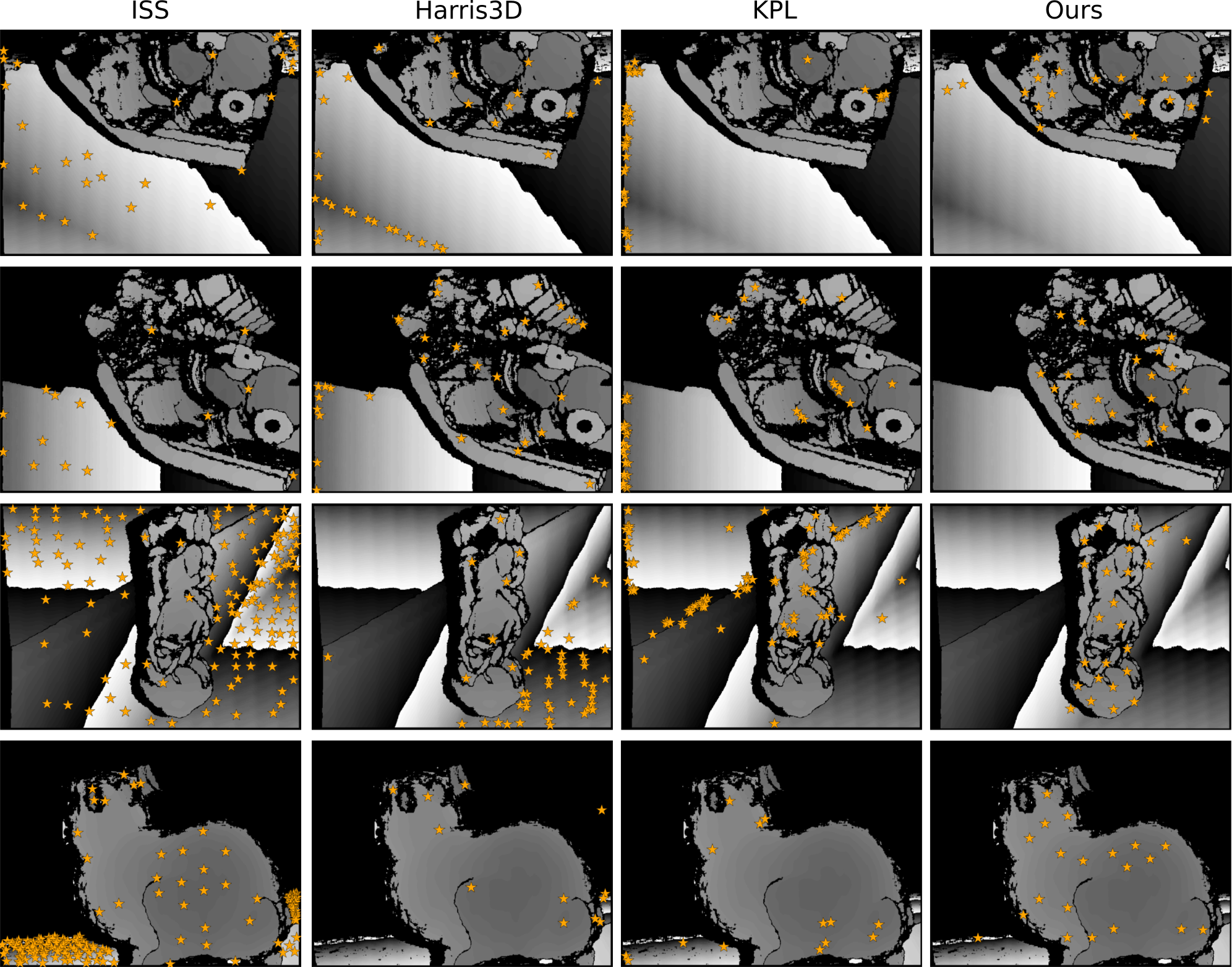}
\end{center}
   \caption{Qualitative evaluation of keypoint generation on the noisy views. Each column represents a different approach. From left to right we have ISS, Harris3D, KPL, and Ours. Notice that the first three methods frequently generate keypoints on background noise, in contrast to our method which generates keypoints mostly on the object. Best viewed in color.}
\label{fig:keypoints_noisy}
\end{figure*}

\begin{figure*}[t]
\begin{center}
 	\includegraphics[width=0.85\linewidth] 
 {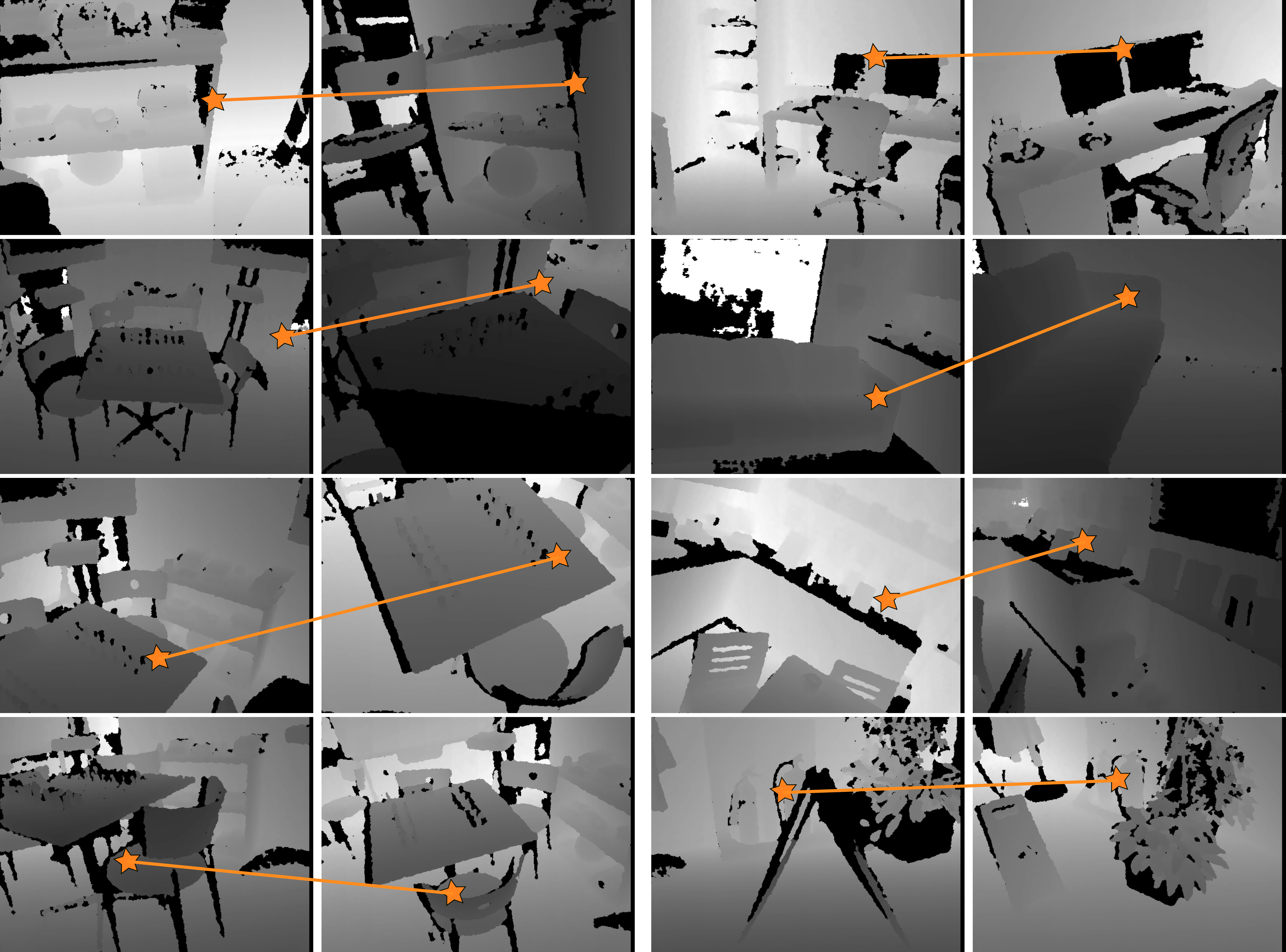}		
\end{center}
	\caption{Keypoint matching examples on the MSR-7 scenes. Columns 1 and 3 show test images and columns 2 and 4 show their retrievals from the repository of descriptors.}   
\label{fig:retrievals_MSR7}
\end{figure*}

\subsection{3D Models}
For this experiment, we use a set of five 3D models, four taken from the Stanford 3D scanning repository~\footnote{http://graphics.stanford.edu/data/3Dscanrep/} (\textit{Armadillo}, \textit{Bunny}, \textit{Dragon}, \textit{Buddha}) and the Honda CBX1000 engine CAD model, from now on referred to simply as \textit{Engine}. Initially, for each 3D model we randomly generate a large number of noise-free views as rendered from the model.
The views are grouped into pairs by first simulating a camera from a certain viewpoint, and then by adding some pose perturbation in order to generate a pair image with some overlap to the first. We use around 10000 image pairs and sample 50 keypoints per image for training each model. For each view, we add simulated depth sensor noise using DepthSynth~\cite{planche_arXiv2017}. The resulting depth images offer much more challenges as noise is present not only on the parts of the 3D model but on its background as well. An example of a pair of views, both noise-free and noisy, can be seen in Figure~\ref{fig:model}.\\
\textbf{Testing protocol.} First, separate training and testing sets of views are generated. After we train our model, a subset of the training set (500 views) is used to generate a repository of descriptors, each assigned to a 3D coordinate. Specifically, we pass each view through our model, collect the descriptors at the predicted keypoint locations, and then project those locations in world coordinates. Then, we apply our model on each view from the test set and match the collected descriptors to the repository. For each descriptor, its nearest neighbour is retrieved. When deciding whether this is a true match, we use a small 3D distance threshold (5 cm) on the distance between the 3D location of the descriptor and its retrieval, and increment the number of true matches accordingly. The reported number is the number of true matches towards the total number of matches. An overview of this procedure is shown in Figure~\ref{fig:pipeline}. Note that we do not use any threshold on the descriptor distance to obtain the set of matches. The same testing procedure is also used for the baselines. For fairness, we tried to keep roughly the same number of generated keypoints per method and per view. \\
\textbf{Engine 3D model.} We investigate the performance of the baselines and our approach on both noise-free and noisy views from the \textit{Engine} model. For this particular experiment we add two more baselines, \textit{Ours-Transfer} and \textit{Ours-Rnd}. For \textit{Ours-Transfer} we train a model on noise-free views, and then test it on the noisy data with the purpose of investigating how well our model can transfer between the noise-free and noisy domains. \textit{Ours-Rnd} randomly selects keypoints instead of using those with the highest scores during the testing procedure. Table~\ref{tab:res_engine} presents the matching accuracies. For the noise-free case, \textit{Ours} is outperformed only from the combination of \textit{Harris3D+FPFH}, which outperforms the deep learning based method of \textit{Harris3D+3DMatch} as well. This is not surprising as these approaches are specifically designed to operate in clean point clouds, however, that is not a realistic setting. Even so, \textit{Ours} demonstrates higher matching accuracy than all the rest of the baselines. 


For the noisy case, we first notice a significant drop in performance from all approaches compared to the noise-free evaluation. \textit{Ours} is the best performing approach, with a difference of $3.1\%$, $6\%$ and $11\%$ towards \textit{Harris3D+3DMatch}, \textit{Ours-Transfer} and \textit{Harris3D+FPFH} respectively. The relatively small gap between \textit{Ours} and \textit{Ours-Transfer} suggests that our model learns to generate good keypoints regardless of the domain it is applied. It is also important to note the large difference between the \textit{Ours-Rnd} and \textit{Ours-No-Score} baselines to \textit{Ours}, which suggests the importance of the score loss during training. Additional qualitative examples are presented in Figure~\ref{fig:no_score_matching} to support our argument, where matching examples are compared between the two methods. After visually examining the examples, we notice that \textit{Ours} produces higher quality matches, most likely due to the consistency of the generated keypoints learned by the score loss.\\
\textbf{Simulated depth sensor noisy views.} Here, we use the 3D models from the Stanford repository and evaluate on their noisy depth images. The testing protocol described earlier is followed, except that we change the 3D distance threshold to 10 cm to account for the errors in the projections of the points. Results shown in Table~\ref{tab:res_synthetic_noisy} follow the same trend as in the \textit{Engine-noisy} evaluation. 
\textit{Ours} is the top performing method, outperforming the next-best baselines \textit{Harris3D+3DMatch} by $13.8\%$ and \textit{Harris3D+FPFH} by $18.2\%$. 
Both combinations with \textit{ISS} fail to retrieve almost any true matches, as \textit{ISS} seems to be the keypoint detector most affected by the simulated sensor noise.
This is a particularly challenging setting for approaches that do not have mechanisms to avoid background noise when generating the keypoints. A qualitative evaluation of the keypoints shown in Figure~\ref{fig:keypoints_noisy} reveals the tendency of the other methods to generate keypoints on noise, while \textit{Ours} focuses on the object. This demonstrates that our method is much less susceptible to the depth sensor noise, and validates our claim for learning the keypoint generation process jointly with the representation.

\begin{table}[t]
\begin{center}
	\begin{tabular}{|c|c|}
	\hline
	Method & Accuracy \\
	\hline
	ISS~\cite{zhong_ICCV2009}+SHOT~\cite{salti2014} & 23.0 \\
	\hline
	ISS~\cite{zhong_ICCV2009}+FPFH~\cite{rusu_ICRA2009} & 24.3\\
	\hline
	Harris3D~\cite{sipiran2011}+FPFH~\cite{rusu_ICRA2009} & 37.4\\
	\hline
	Harris3D~\cite{sipiran2011}+SHOT~\cite{salti2014} & 37.9 \\
	\hline
	Harris3D~\cite{sipiran2011}+3DMatch~\cite{zeng20173dmatch} & 38.2 \\
	\hline
	Ours & \textbf{41.2} \\
	\hline
  	\end{tabular}
    \caption{Keypoint matching accuracies (\%) on the MSR-7~\cite{shotton_CVPR2013} dataset.}
    \vspace{-1cm}
    \label{tab:res_msr7}
\end{center}
\end{table}
\textbf{Computational cost.} Our end-to-end method requires only 0.14s per image to perform a forward pass of the network and generate keypoints and descriptors. For comparison, LIFT~\cite{Yi_ECCV16} takes 2.78s per image on the same machine. During training our method needs 0.4s per iteration. All times reported are on an NVIDIA Titan GPU for the honda engine noisy data with 50 keypoints generated per image.

\subsection{Real Depth Sensor}
\textbf{MSR-7.} For this experiment, we use the publicly available MSR-7 scenes dataset~\cite{shotton_CVPR2013}, which offers RGB-D sequences captured with Kinect and reconstructions of indoor scenes. We followed the train-test sequence split provided, and trained a model for each scene on 10-frame-apart pairs of the depth images. The same testing protocol of keypoint matching as in the previous experiments is employed. We do not use the baseline \textit{Ours-No-Score} as it consistently underperformed in the previous experiments, nor the \textit{KPL} because it was trained on a very different dataset.

\begin{figure}[t]
\begin{center}
 	\includegraphics[width=1\linewidth]{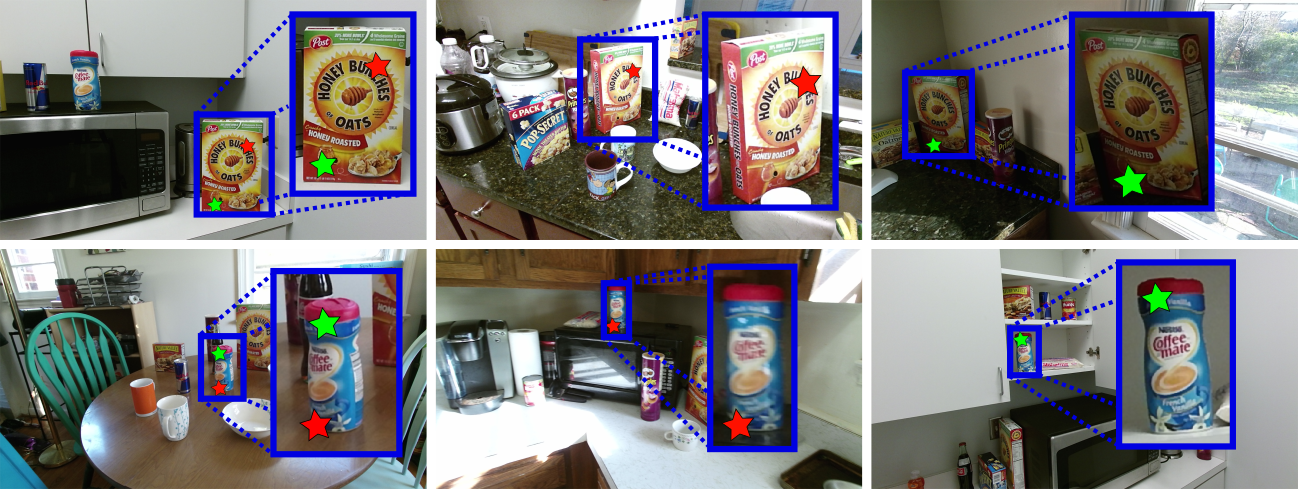}	
\end{center}
   \vspace{-1em}
   \caption{Matching examples from GMU-Kitchens. First column shows queries and retrieved points are color-coded (zoomed-in for clarity). Note that we use the depth map for our experiments but we show the retrievals in RGB for
the sake of clarity.}
\vspace{-1.5em}
\label{fig:retrievals_gmu}
\end{figure}

Table~\ref{tab:res_msr7} shows the average matching accuracy over all scenes. Again, our method seems to have the edge over the baselines, with a $3\%$ improvement over the second-best \textit{Harris3D+3DMatch}. This result suggests that our approach can be successfully applied on sequences captured by a real sensor, besides 3D models with simulated noise.
In Figure~\ref{fig:retrievals_MSR7} we present some retrieval examples from different scenes in the dataset. We make a similar observation as in the noise-free experiment, where true matches were retrieved from larger viewpoint variations than the ones provided during training. Note that the training pairs, 10-frames-apart, have small pose differences.\\
\textbf{GMU-Kitchens} In this experiment we qualitatively investigate the performance of our method on objects captured by Kinect-v2. We use the publicly available GMU-Kitchens~\cite{georgakis_3DV2016} dataset which contains 9 RGB-D videos of kitchen scenes with 11 object instances from the BigBIRD~\cite{singh_ICRA2014} dataset. Unlike the previous experiment where we generate keypoints in the scenes, here we focus on matching keypoints generated inside the bounding boxes of objects. In particular, we use the train-test split of fold 1 as defined in~\cite{georgakis_3DV2016}, and for each object we create a repository of descriptors. Then, the descriptors collected from the bounding boxes in the test scenes are matched to the appropriate object repository (see Figure~\ref{fig:retrievals_gmu}). Note that the model was trained by sampling keypoints from the depth maps in the scenes, similar to the MSR-7 experiment, and not specifically from the object bounding boxes. 

\vspace{-0.2cm}
\section{Conclusions}
We presented a unified, end-to-end, framework to simultaneously learn a keypoint detector and view-invariant representations of keypoints for 3D keypoint matching. To learn view-invariant representations, we presented a novel sampling layer that creates ground-truth data on-the-fly, generating pairs of keypoint proposals that we use to optimize a constrastive loss objective function. Furthermore, to learn to generate the right keypoint proposals from a keypoint matching perspective, we introduced a new score loss objective that maximizes the number of positive matches between images from two viewpoints. We conducted keypoint matching experiments on multiple 3D benchmark datasets and demonstrated qualitative and quantitative improvements over the existing state-of-the-art. 

{\small
\bibliographystyle{ieee}
\bibliography{egbib}
}

\end{document}